\documentclass[10pt,twocolumn,letterpaper]{article}

\usepackage{cvpr}
\usepackage{epsfig}
\usepackage{graphicx}
\usepackage{amsmath}
\usepackage{amssymb}
\usepackage{dsfont}
\usepackage{booktabs}
\usepackage{makecell}
\usepackage{float}
\usepackage{diagbox,tabularx}

\usepackage[font=small]{caption}

\newcommand{\similarity}{s}
\newcommand{\posa}{P_1}
\newcommand{\posb}{P_2}
\newcommand{\negai}{N_i}

\newcommand{\loss}{L}

\usepackage{authblk}

\cvprfinalcopy 

\def\cvprPaperID{4295} 

\ifcvprfinal\fi
\begin{document}
	
	\title{Discovering Visual Patterns in Art Collections \\
		with Spatially-consistent Feature Learning}
	
	\author[1]{Xi Shen}
	\author[2]{Alexei A. Efros}
	\author[3]{Mathieu Aubry}
	
	\affil[1,3]{LIGM (UMR 8049), \'Ecole des Ponts ParisTech}
	\affil[2]{University of California, Berkeley}
	
	\makeatletter
	\renewcommand\AB@affilsepx{, \protect\Affilfont}
	\makeatother
	\affil[1,3]{\{xi.shen, mathieu.aubry\}@enpc.fr}
	\affil[2]{efros@eecs.berkeley.edu}

	\maketitle
	
	\begin{abstract}
		\vspace{-2mm}
		Our goal in this paper is to discover near duplicate patterns in large collections of artworks. This is harder than standard instance mining due to differences in the artistic media (oil, pastel, drawing, etc), and imperfections inherent in the copying process.
		Our key technical insight is to adapt a standard deep feature to this task by fine-tuning it on the specific art collection using self-supervised learning.  More specifically, spatial consistency between neighbouring feature matches is used as supervisory fine-tuning signal.
		The adapted feature leads to more accurate style-invariant matching, and can be used with a standard discovery approach, based on geometric verification, to identify duplicate patterns in the dataset. 
		The approach is evaluated on several different datasets and shows surprisingly good qualitative discovery results. For quantitative evaluation of the method, we annotated 273 near duplicate details in a dataset of 1587 artworks attributed to Jan Brueghel and his workshop\footnote{Code and data is available at \url{http://imagine.enpc.fr/~shenx/ArtMiner}}. Beyond artworks, we also demonstrate improvement on localization on the Oxford5K photo dataset as well as on historical photograph localization on the Large Time Lags Location (LTLL) dataset.
		
	\end{abstract}

	\begin{figure}[!t]
		\centering
		\begin{minipage}{\linewidth}
			\centering
			\includegraphics[width=0.32\linewidth, height=0.32\linewidth]{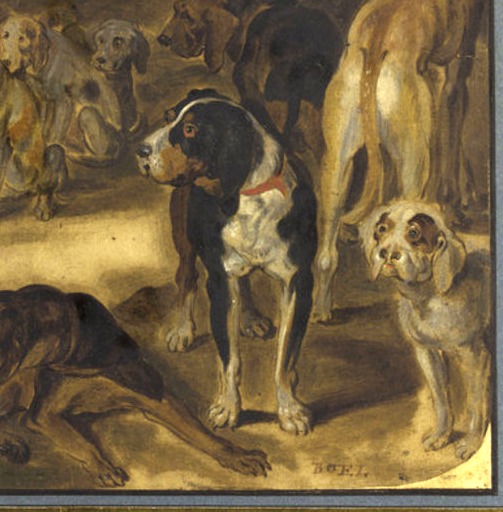}
			\includegraphics[width=0.32\linewidth, height=0.32\linewidth]{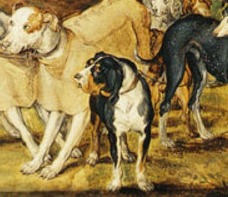}
			\includegraphics[width=0.32\linewidth, height=0.32\linewidth]{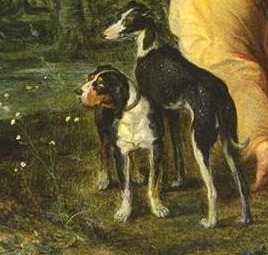}
			\vspace{-.1in}
			\caption*{(a)  common detail (dog) discovered in our new Brueghel dataset of 1587 artworks}
		\end{minipage}
		\begin{minipage}{0.49\textwidth}
			\centering
			\includegraphics[width=\linewidth]{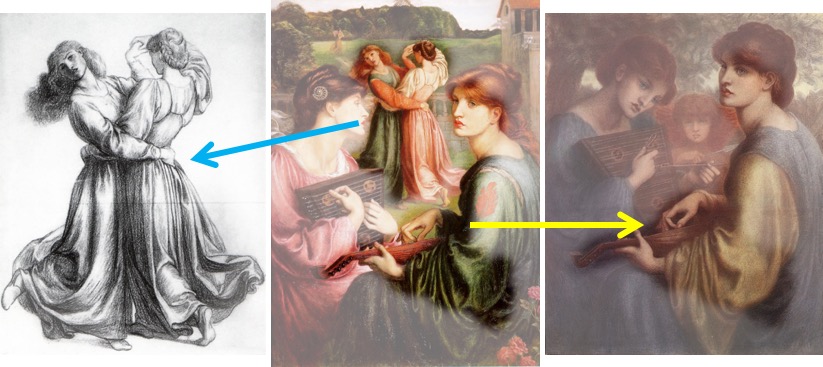}
			\vspace{-.25in}
			\caption*{(b) relationship between painting and two studies discovered from collection of 195 artworks by Dante Gabriel Rossetti}
		\end{minipage}
		\vspace{-.1in}
		\caption{Examples of repeated visual patterns automatically discovered by our algorithm. Sources: (a) left: {\em Nymphs Sleeping After the Hunt, Spied on by Satyr} (oil), center: {\em Diana's Nymphs After the Hunt} (oil), right: {\em Seventeen Studies of Different Dogs} (drawing); (b) {\em The Bower Meadow} (left: chalk, center: oil, right: pastel)}
		
		\label{fig:teaser}
		\vspace{-5mm}
	\end{figure}
	
	\section{Introduction}

	Visiting a world-class art museum might leave one with an impression that each painting is absolutely unique and unlike any other. In reality, things are more complicated.  While working on a painting, an artist would typically create a number of preliminary sketches and studies to experiment with various aspects of the composition.  Many of these studies also find their way into (usually more provincial) museums. Some artists enjoy returning time and again to the same subject matter (e.g. Claude Monet and his series of 25 paintings of the same haystacks).  Moreover, during the Renaissance, it was not uncommon for an artist (or an apprentice in his workshop) to reuse the same visual elements or motifs (an angel, a cart, a windmill, etc) in multiple paintings, with little or no variation.  For example, Flemish painter Jan Brueghel is believed to have created many paintings that were imitations, pastiches, or reworkings of his own works, as well as these of his father, Pieter Breughel the Elder~\cite{honig2016jan}.  Art historians are keenly interested in mapping out such visual connections between artworks to discover influences, find provenance, and even establish authorship.   Currently, this is being done entirely by hand, with researchers spending months or even years in museum archives hoping to discover common visual patterns.

	This paper presents an approach for automatically discovering repeated visual patterns in art collections, as shown on Figure~\ref{fig:teaser}.
	We propose to learn a deep visual feature able to find correspondences between near-duplicate visual elements across different artworks. 
	This task is quite challenging, requiring a feature  that is both highly discriminative (i.e. tuned to find copies of the same object instance rather then samples of an object category), but also invariant to changes in color, style, artistic media, geometric deformation, etc.   Manually collecting and labelling a large enough artwork dataset containing enough variability would require enormous effort by professional art historians, which is exactly what we are trying to avoid.  Therefore, we propose a method which learns in a self-supervised way, adapting a deep feature to a given art collection without any human labelling. 
	This is done by leveraging neighbourhood spatial consistency across matches as free supervisory signal.

	Using our trained feature, we demonstrate that a simple voting and regression procedure, in line with classic verification step of instance recognition ~\cite{philbin2007object}, lets us discover visual patterns that are repeated across artworks within the dataset. 
	
	We demonstrate our visual pattern discovery approach on several collections of artwork, including a new annotated dataset of 1587 works attributed to the 
	Brueghel family.   To further evaluate the generality of our method, we have also used it to perform localization by matching a set of reference photographs to a test image. We demonstrate that our feature learning procedure provides a clear improvement on the Oxford5K dataset~\cite{philbin2007object}, and results in state-of-the-art performance on the Large Time Gap Location dataset~\cite{fernando2015location} for localizing historical architecture photographs with respect to modern ones.
	
	Our main contributions are: 1) a self-supervised approach to learn a feature for matching artistic visual content across wide range of styles, media, etc; 2) the introduction of a large new dataset for evaluating visual correspondence matching; 3) an approach to discover automatically repeated elements in artwork collections.
	

	\section{Related Work}

	\paragraph{Computer vision and art. }
	There is a long standing and fruitful collaboration between computer vision and art.   On the synthesis side, promising results have been obtained for transferring artistic style to a photograph ~\cite{hertzmann2001image,gatys2016image,zhu2017unpaired}, or even trying to create art ~\cite{elgammal2017can,hertzmann2018can}. 
	On the analysis side, there are several efforts on collection and annotation of large-scale art datasets~\cite{karayev2013recognizing, mensink2014rijksmuseum,picard2015challenges, wilber2017bam,strezoski2017omniart, mao2017deepart},
	and using them for genre and authorship classification~\cite{karayev2013recognizing,tan2016ceci,strezoski2017omniart}. 
	Others focus on applying and generalizing visual correspondence and object detection methods to paintings using both classical ~\cite{shrivastava2011, ginosar2014detecting,Crowley13,aubry2014painting}, as well as deep~\cite{crowley2015face,crowley2016art,westlake2016detecting,gonthier2018weakly} methods.  Most closely related to us is work of Yin et al~\cite{yin2016object}, which used the same  Brueghel data~\cite{brueghel}, annotating it to train detectors for five object categories (carts, cows, windmills, rowboats and sailbaots).

	Our goal, however, is to go further and focus on the computational analysis of relationships {\em between} individual artworks. Seguin et al~\cite{seguin2016visual,seguin2017tracking}  propose to find visual relationships in collections of paintings. However, while they use off-the-shelf CNNs trained in a supervised manner, we focus on the design of a new self-supervised feature learning specifically trained for the task.
	This allows us to focus on near-exact reproduction of detail, rather than a more generic visual similarity, which is what most art historians are actually interested in regards to specific corpora, such as the works of Brueghel family~\cite{honig2016jan}.
	

	
	

	\begin{figure*}[t!]
		\begin{center}
			\begin{minipage}{0.32\textwidth}
				\centering    
				\includegraphics[height=3.5cm]{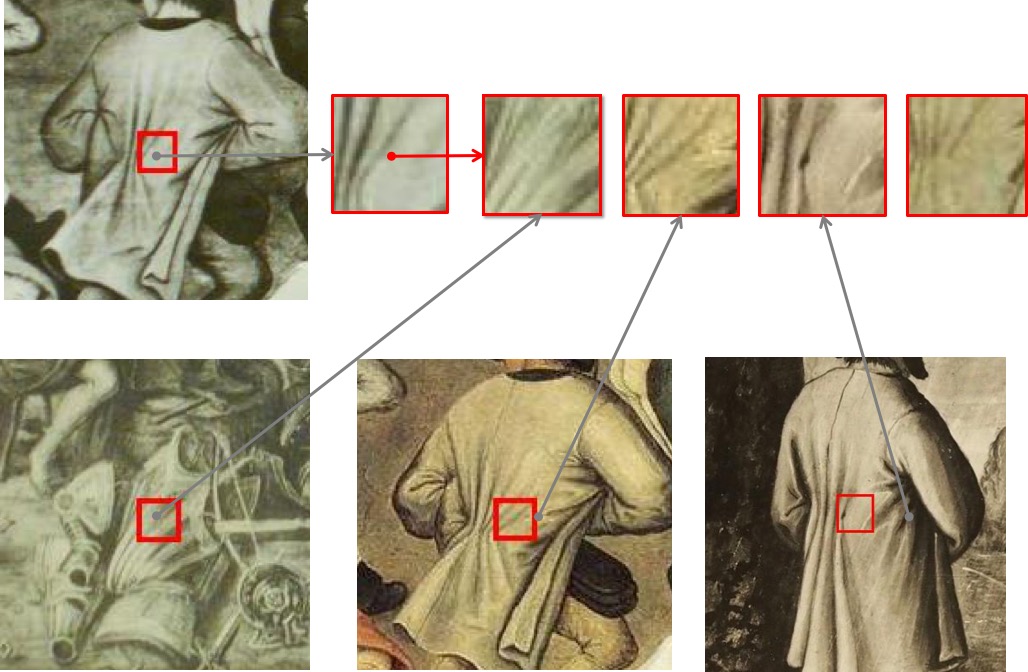}
				\vspace{-3mm}
				\caption*{(a) Candidates from proposal region}
			\end{minipage}~\vline~
			\begin{minipage}{0.5\textwidth}
				\centering
				\includegraphics[height=3.5cm]{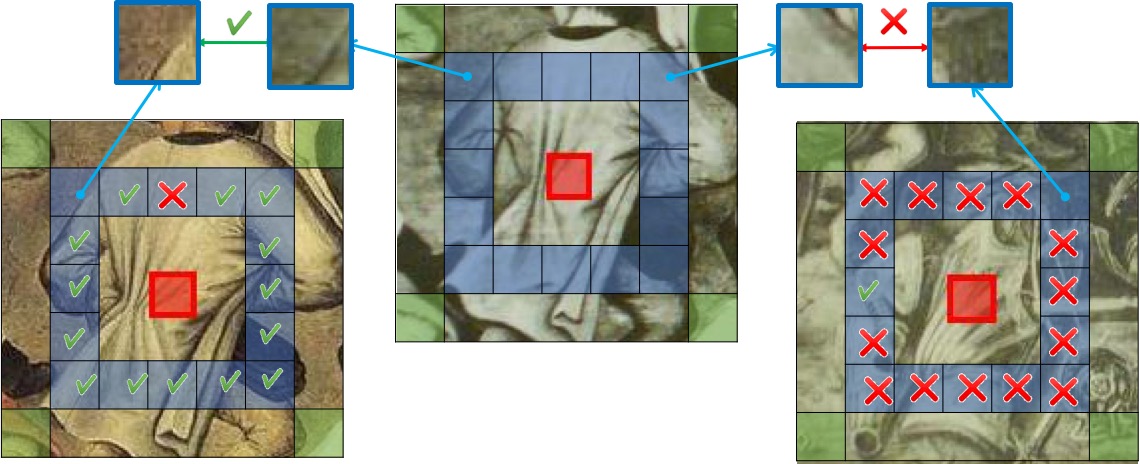}
				\caption*{(b) Selection with verification regions}
			\end{minipage}\vline
			\begin{minipage}{0.17\textwidth}
				\centering
				~\includegraphics[height=3.cm]{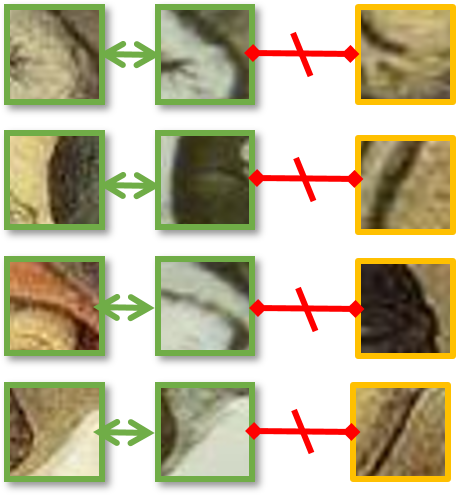}
				\caption*{\centering (c) Training with positive regions}
			\end{minipage}
		\end{center}
		\vspace{-7mm}
		\caption{Feature Learning Strategy. (a) Our approach relies on candidate correspondences obtained by matching the features of a proposal region (in red) to the full database . (b) The candidate correspondences are then verified by matching the features of the verification region (in blue) of the query in the candidate images and checking for consistency . (c) Finally, we extract features from the positive regions (in green) from the verified candidates and use them to improve the features using a metric learning loss. }
		\label{fig:feature_learning}
		\vspace{-5mm}
	\end{figure*}

	\vspace{-3mm}
	\paragraph{Spatial consistency as supervisory signal.}
	Spatial consistency is a widely used signal in many computer vision tasks from geometry to retrieval. 
	The classic work of Sivic et al.~\cite{sivic2003video} performs instance retrieval based on the extraction of spatially consistent local feature matches. This direction has been further developed with specially adapted features for place recognition across large visual changes ~\cite{fernando2015location,hauagge2012image,verdie2015tilde, aubry2014painting}. Beyond instances, this idea has been extended to discovering object categories~\cite{cho2015unsupervised} and their segmentations~\cite{rubinstein2013unsupervised}.
	Our discovery of repeated patterns through correspondence consistency is reminiscent of the line of work on mid-level visual element discovery ~\cite{singh2012unsupervised,doersch2013mid,doersch2014context}. These idea have been used in the context of temporal and spatial image collection analysis, to discover the elements characteristic of a specific location~\cite{doersch2012makes}, or the evolution of these elements over time~\cite{jae2013style}.
	
	Spatial consistency has also been used to learn deep visual features for object category in a self-supervised way, either by predicting the spatial configuration of patches ~\cite{doersch2015unsupervised} or predicting the patch given its context~\cite{pathak2016context}. 
	In a similar spirit, Rocco et al.~\cite{rocco2018neighbourhood} recently demonstrated how to learn visual representations through geometric consistency to predict object-category-level correspondences between images.  We, on the other hand, aim at learning features for matching only stylistically different versions of the same instance.

	\section{Dataset-specific Feature Learning}

	This section describes our strategy for adapting deep features to the task of matching artworks across styles in a specific dataset.  Starting with standard ImageNet pre-trained deep features, our idea is to extract hard-positive matching regions form the dataset that we then use in a metric learning approach to improve the features. Our two key hypothesis are that: (i) our dataset includes large parts of images that are copied from each other but are depicted with different styles, and (ii) the initial feature descriptor is good enough to extract some positive matches. 
	Our training thus alternates between two steps that we describe bellow: (1) mining for hard-positive training samples in the dataset based on the current features using spatial consistency, and (2) updating the features by performing a single gradient step on the selected samples.

	\subsection{Mining for Positive Feature Pairs}
	\label{sec:mpp}

	For our approach to work, it is crucial to select positive matching examples that are both accurate and difficult. Indeed, if the features are trained with false matches, training will quickly diverge, and if the matches are too easy, no progress will be made.
	
	To find these hard-positive matching features, we rely on the procedure visualized in Figure~\ref{fig:feature_learning}. 
	
	\vspace{-3mm}
	\paragraph{Candidate sampling.} 
	
	{\it Proposal regions} are randomly sampled from each image in the dataset to be used as query features. These are matched densely at every scale to all the images in the dataset using cosine similarity in feature space. This can be done efficiently and in parallel for many queries using a normalization and a convolution layer, with the weights of the convolution defined by the query features.
	For each query we select one of its top $K$ matches 
	as candidate correspondences (Figure~\ref{fig:feature_learning}a). 
	These candidates contain a high proportion of bad matches, since most of the queries are likely not repeated $K$ times in the dataset and since our feature is imperfect.



\begin{figure}[!t]
		\begin{center}
			\begin{minipage}{0.057\textwidth}
				\centering    
				\includegraphics[width=\textwidth, height=\textwidth]{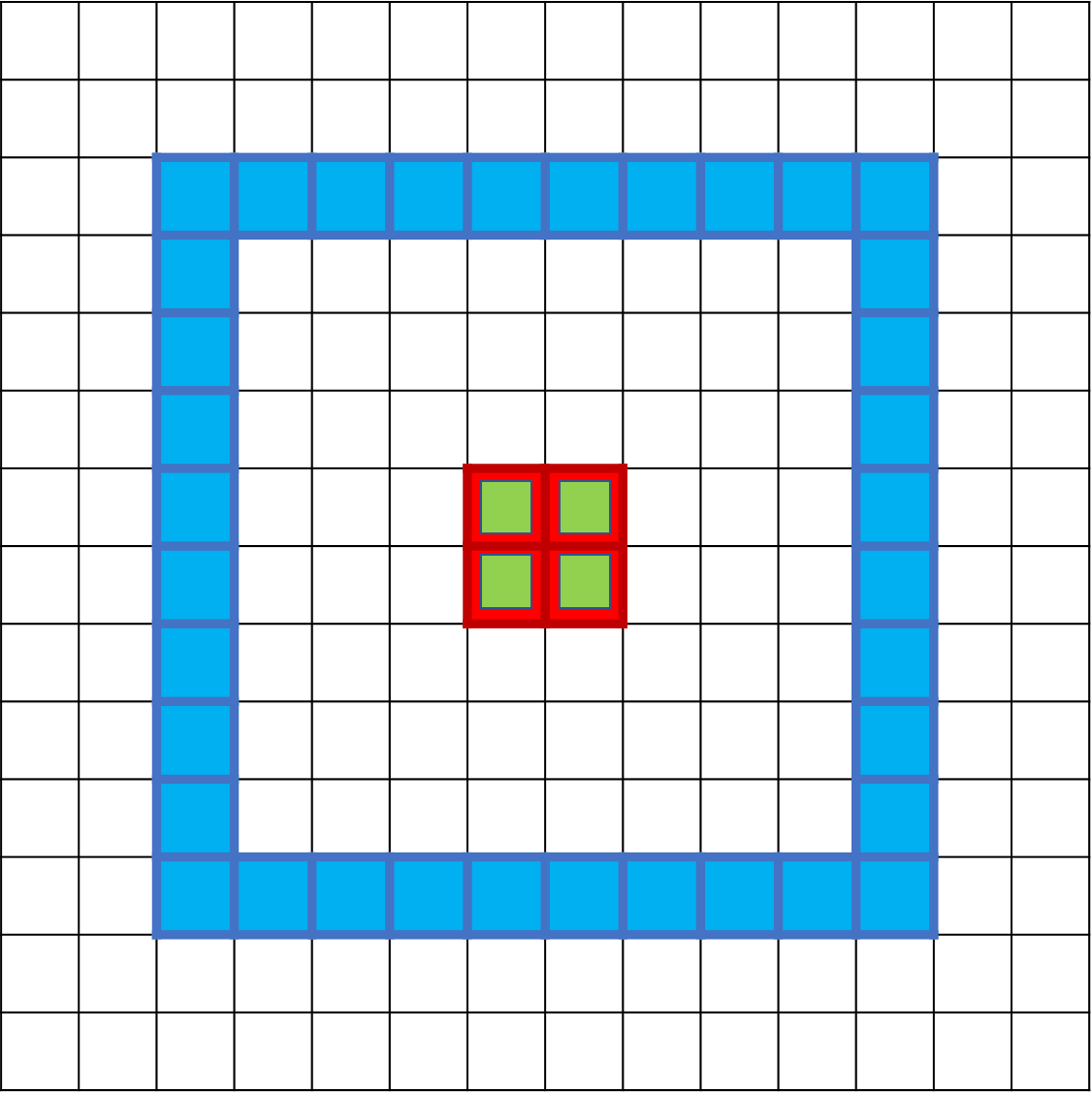}
				\vspace{-7mm}
				\caption*{P2}
				
			\end{minipage}
			\hspace{0.01cm}
			\begin{minipage}{0.057\textwidth}
				\centering    
				\includegraphics[width=\textwidth, height=\textwidth]{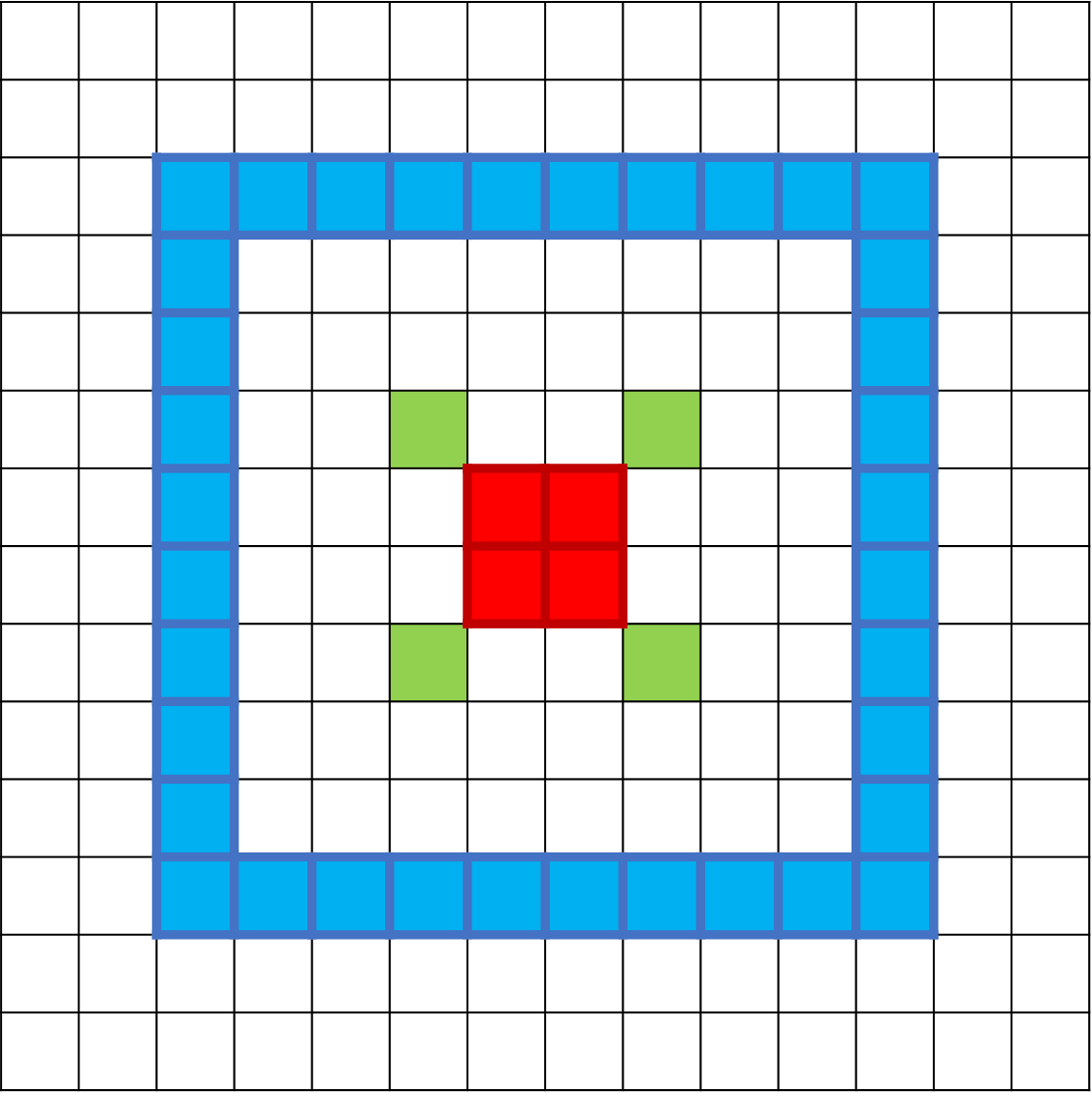}
				\vspace{-7mm}
				\caption*{P4}
			\end{minipage}
			\hspace{0.01cm}
			\begin{minipage}{0.057\textwidth}
				\centering    
				\includegraphics[width=\textwidth, height=\textwidth]{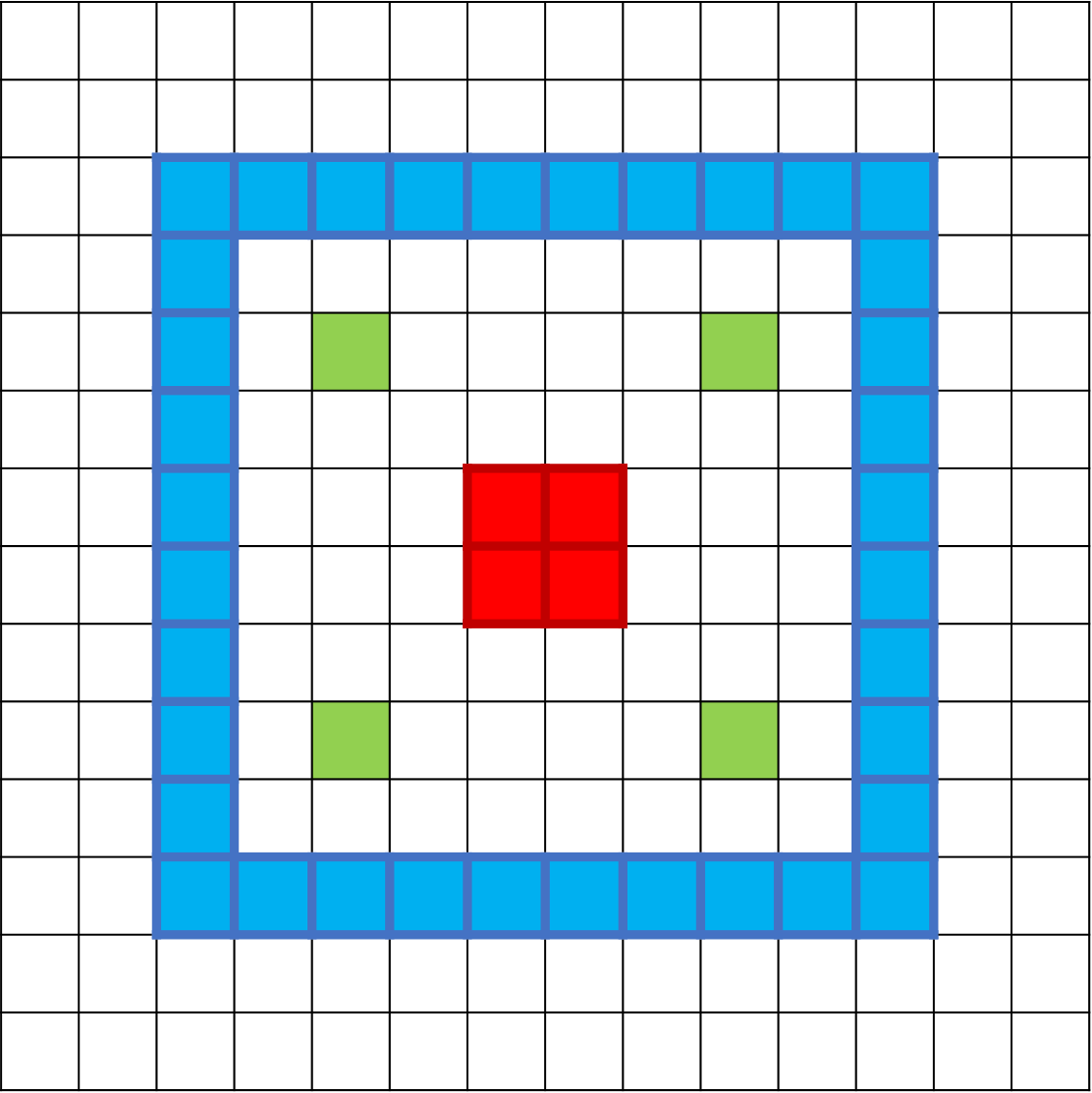}
				\vspace{-7mm}
				\caption*{P6}
			\end{minipage}
			\hspace{0.01cm}
			\begin{minipage}{0.057\textwidth}
				\centering    
				\includegraphics[width=\textwidth, height=\textwidth]{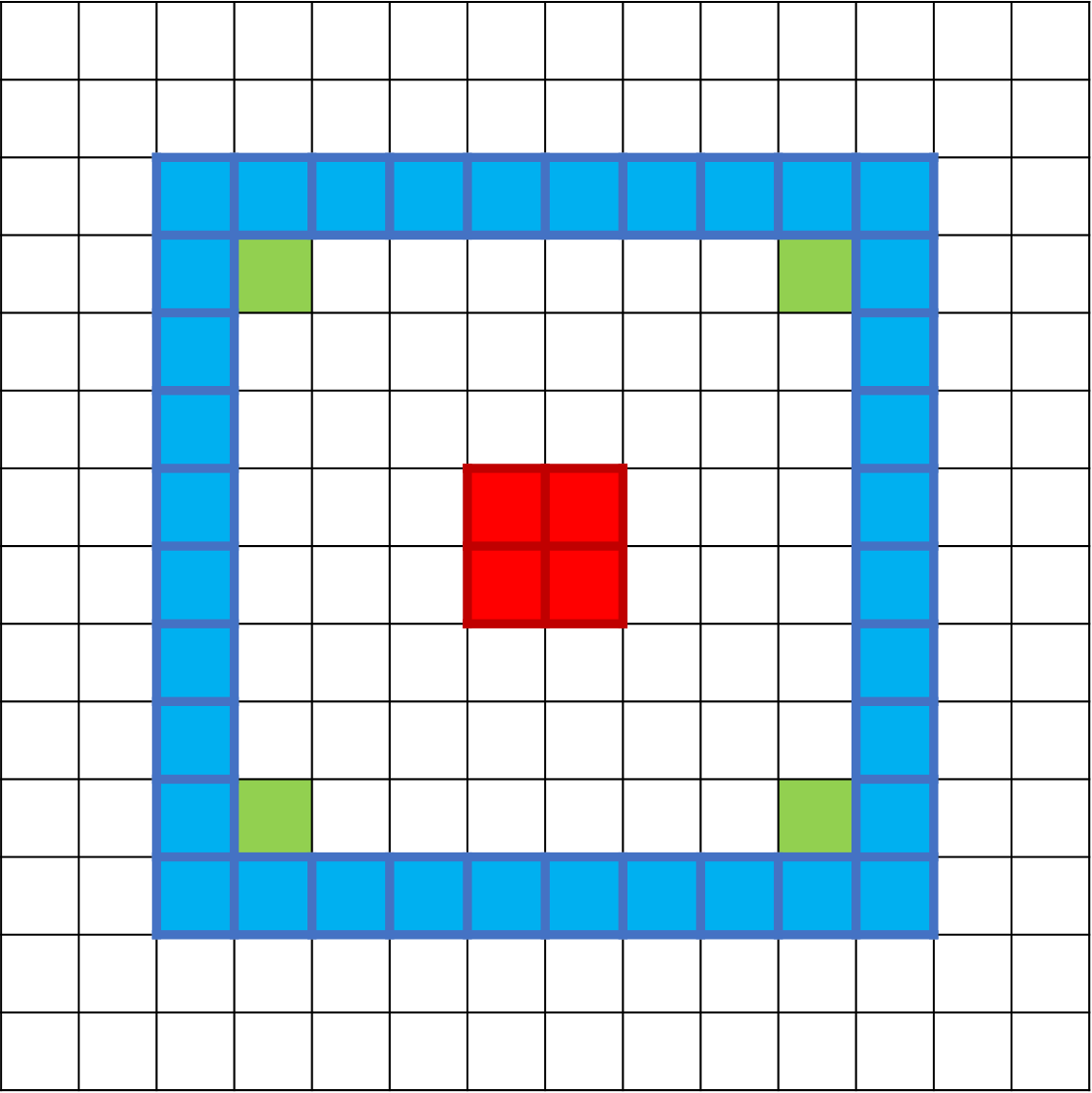}
				\vspace{-7mm}
				\caption*{P8}
			\end{minipage}
			\hspace{0.01cm}
			\begin{minipage}{0.057\textwidth}
				\centering    
				\includegraphics[width=\textwidth, height=\textwidth]{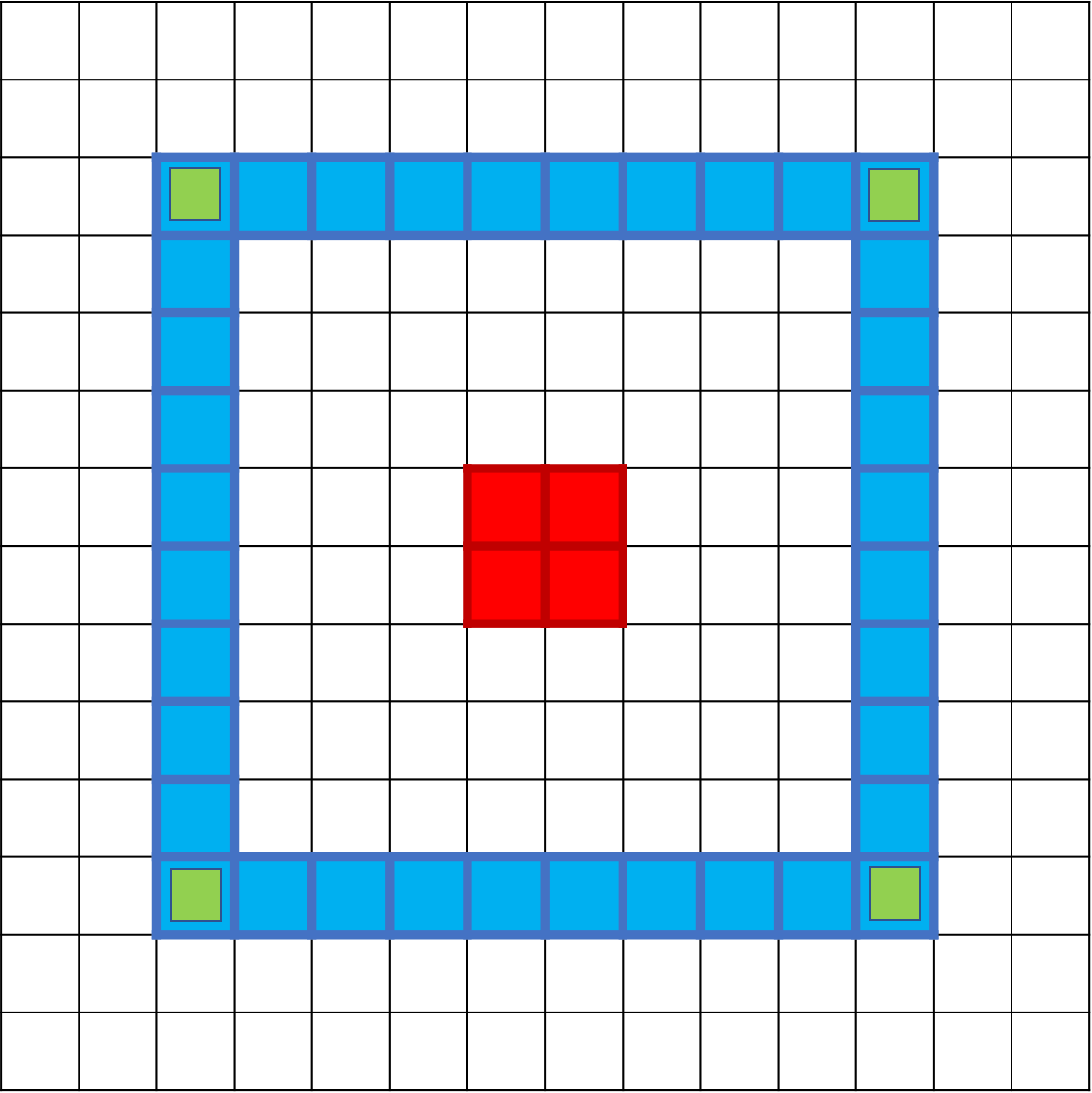}
				\vspace{-7mm}
				\caption*{P10}
			\end{minipage}
			\hspace{0.01cm}
			\begin{minipage}{0.057\textwidth}
				\centering    
				\includegraphics[width=\textwidth, height=\textwidth]{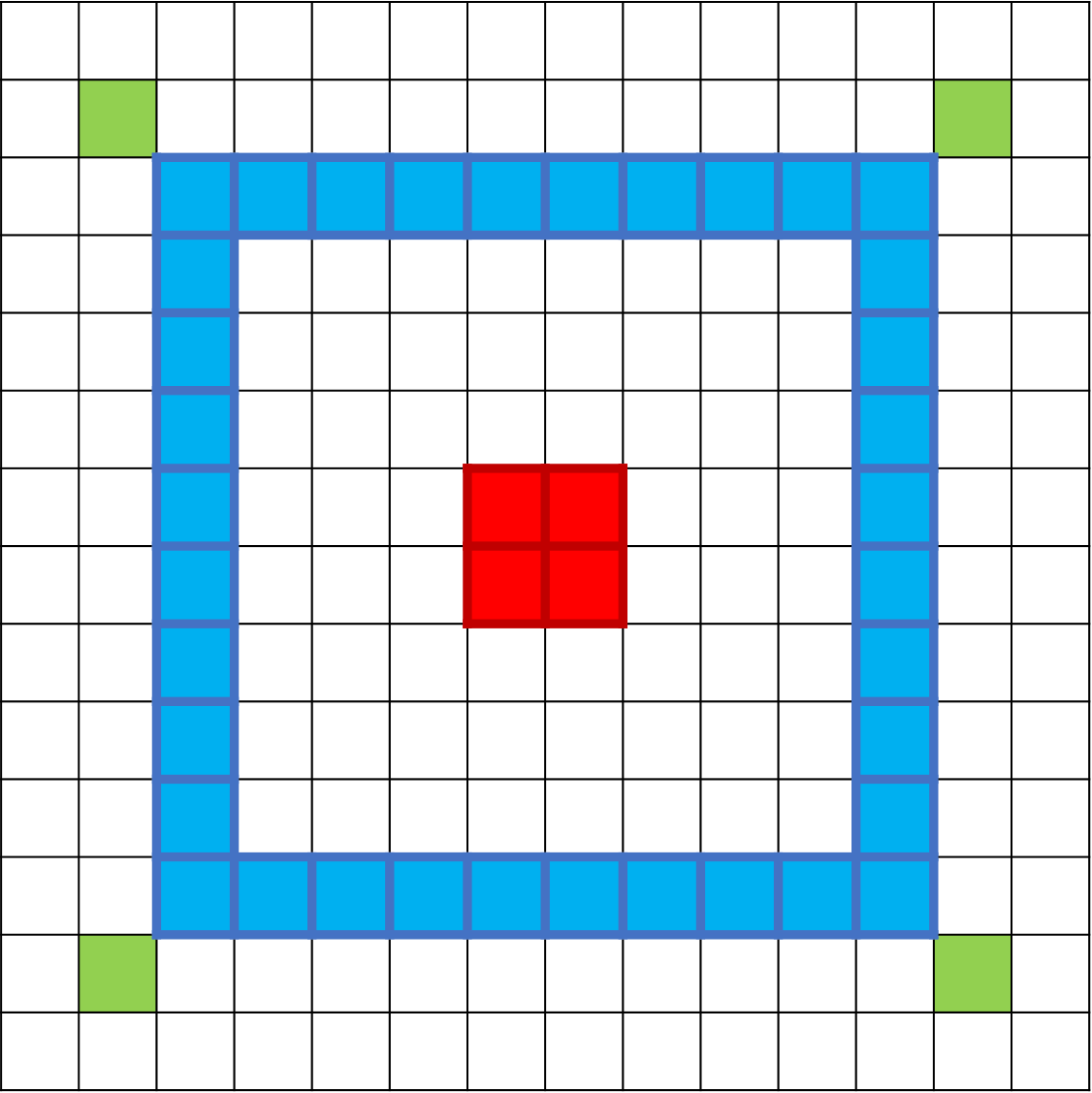}
				\vspace{-7mm}
				\caption*{P12}
			\end{minipage}
			\hspace{0.01cm}
			\begin{minipage}{0.057\textwidth}
				\centering    
				\includegraphics[width=\textwidth, height=\textwidth]{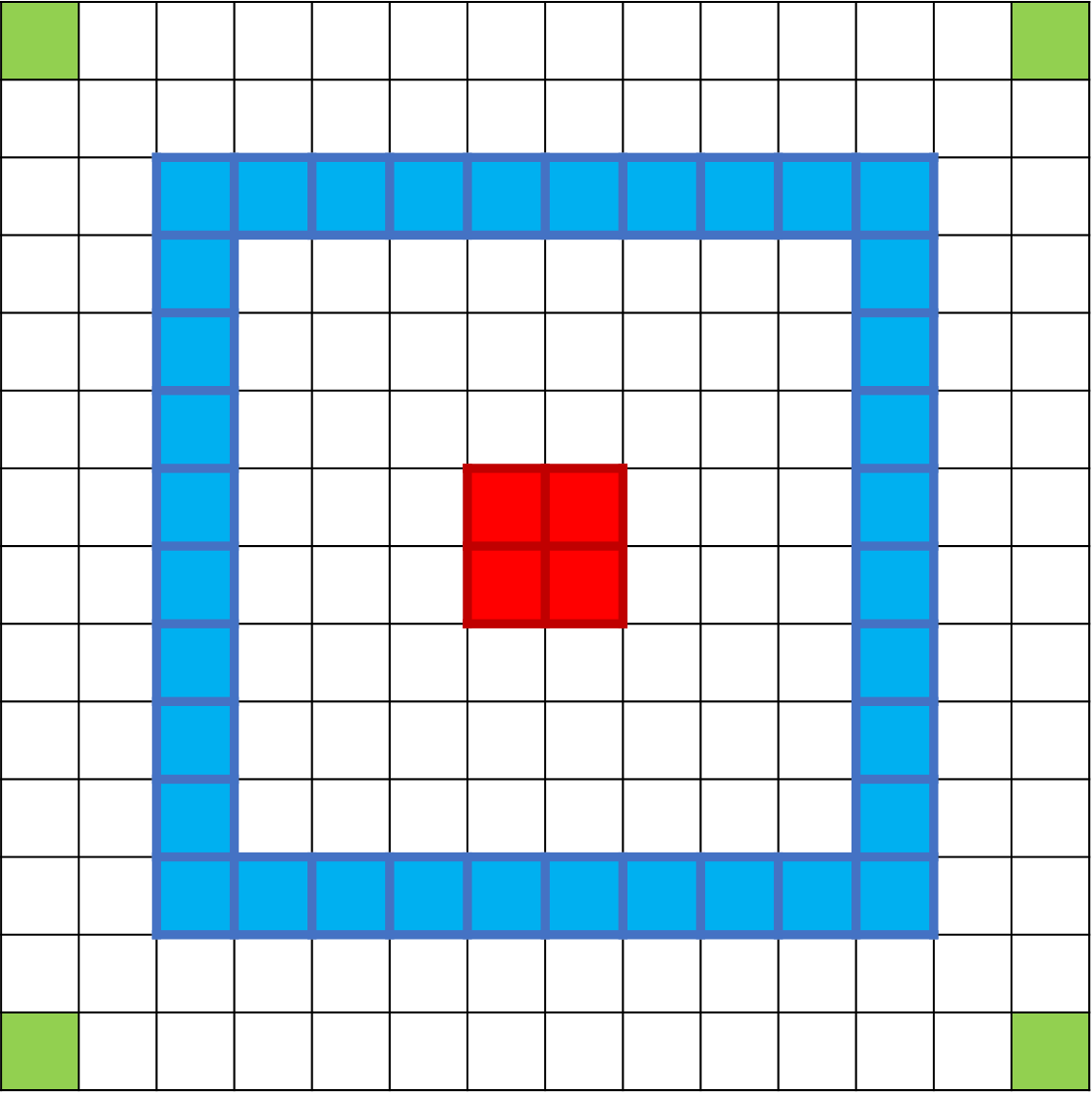}
				\vspace{-7mm}
				\caption*{P14}
			\end{minipage}
			
		\end{center}
		\vspace{-7mm}
		\caption{Different region configurations.
			(red: proposal regions,
			blue: verification regions,
			green: positive regions)}
		\label{fig:train_setting}
		\vspace{-5mm}
	\end{figure}

	\vspace{-3mm}
	\paragraph{Candidate verification.} 
	To verify the quality of candidate matches given by the previous step, we rely on special consistency: a match will be considered valid if its neighbours agree with it. More precisely, let's assume we have a candidate match between features from the proposal region $p_A$ in image $A$ and a corresponding region $p_B$ in image $B$, visualized in red in Figure~\ref{fig:feature_learning}b and~\ref{fig:train_setting}.
	We define a {\it verification region} around $p_A$, visualized in blue. Every feature in this region is individually matched in image B, and votes for the candidate match if it matched consistently with $p_B$. Summing the votes of all the features in the verification region allows us to rank the candidate matches. A fixed percentage of the candidates are then considered verified. 
	The choice of the verification region is, of course, important to the success of this verification step. The key aspect is that the features in the verification region should be, as much as possible, independent of the features in the proposal region. On the other hand, having them too far apart would reduce the chances of the region being completely matched. For our experiments, we used the 10x10 feature square centred around the query region (Figure~\ref{fig:train_setting}). 
	
	\vspace{-3mm}
	\paragraph{Generating hard positives.} Finally, given a set of verified correspondences, we have to decide which features to use as positive training pairs.
	One possibility would be to directly use features in the proposal region, since they have been verified. However, since the proposal region has already been ``used'' once (to verify the matches), it does not bring enough independent signal to make quality hard positives. Instead, we propose to sample positives from a different {\it positive region}.
	We evaluated different configurations for the positive region, as visualized in Figure~\ref{fig:train_setting} (in green). 
	We choose to keep only 4 positive pairs per verified proposal, positioned at the corners of a square and denote the different setups as P2 to P14, the number corresponding to the size of the square. We will show in Appendix~\ref{sec:exp_diff_region} that P12 and P14 perform better than the alternatives. 
	

	\subsection{Feature Fine-tuning} 

	After each selection of positive feature pairs (Figure~\ref{fig:feature_learning}b in green), we update of our feature (Figure~\ref{fig:feature_learning}c) using a single gradient step of the following triplet metric learning loss: 
	
	\vspace{-5mm}
	\begin{eqnarray}
	\loss(\posa, \posb, \{\negai\}) =-\min(\lambda , \similarity(\posa, \posb)) +  \nonumber \\ \frac{1}{N_{neg}} \sum_{i=1}^{N_{neg}} \max(\similarity(\posa, \negai), 1 - \lambda)   
	\label{eqn:triplet-loss}
	\end{eqnarray}
	
	\vspace{-2mm}
	where $\posa$ and $\posb$ are corresponding features in the positive regions, $\{N_i\}_{\{i = 1, 2...N_{neg}\}}$ are negative samples, s is the cosine similarity metric and $\lambda$ is a hyper-parameter.
	We select the negatives as the set of top matches to $\posa$ in $\posb$'s image. This selects hard negatives and avoids any difference in the distribution of the  depiction styles in our positive and negative samples. We chose a relatively high number of negatives $N_{neg}$ to account for the fact that some of them might in fact correspond to matching regions, for example in the case of repeated elements, or for location near the optimal match.

	\vspace{-3mm}
	\paragraph{Implementation details.}
	In all of our experiments, 
	we used {\it conv4} features of the ResNet-18~\cite{he2016deep} architecture. 
	We resized all images such that their maximum spatial dimension in the feature map was 40, leading to approximately 1k features per image at the maximum scale. 
	For each image, we used 7 different scales, regularly sampled on two octaves with 3 scales per octave. 
	For positive sampling, we used square queries of $2\times 2$ features, $K=10$ candidate matches for each query. 
	From these candidate, the top 10\% with the most votes from neighbours were considered verified. Note that these parameters might need to be adjusted depending on the diversity and size of the dataset, but we found that they performed well both for the Brueghel~\cite{brueghel} and LTLL~\cite{fernando2015location} datasets. For training on Oxford5K~\cite{philbin2007object} dataset, the query patches are only sampled inside the annotaed bounding boxes of the 55 query images, and we only find candidate matches in 2000 images randomly sampled in the whole dataset.
	The hyper-parameters of the triplet loss, $N_{neg}$  and $\lambda$, are fixed to 20 and 0.8 respectively. 
	Our models were trained with the Adam~\cite{kingma2014adam} optimizer with learning rate 1e-5 and $\beta = [0.9, 0.99]$. Using a single GPU Geforce GTX 1080 Ti, training converged in approximately 10 hours, corresponding to 200 iterations of the feature selection and training. Most of the time is spent extracting and verifying candidate matches.  ImageNet pre-training was used for initialization in all experiments.
	
 
	\section{Spatially consistent pattern mining}

	\begin{figure}[!t]
		\centering    
		\includegraphics[width=\linewidth]{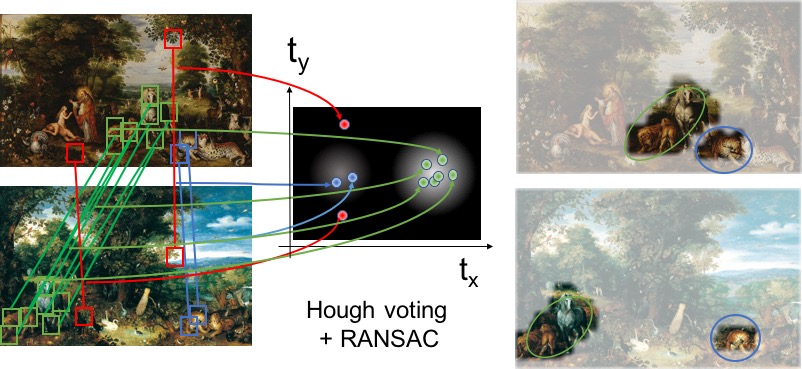}
		\vspace{-7mm}
		\caption{Discovery through geometric consistency. }
		\label{fig:discovery}
		\vspace{-3mm}
	\end{figure}
	
	In this section, we describe how our algorithm discovers repeated patterns in a dataset given style-invariant features learned in the previous section. We follow the classic geometric verification approach~\cite{philbin2007object}: for all pairs of images in the dataset, we first compute features and match them between the two images, then select consistent matches, and finally find image regions that have many consistent feature matches. This allows us to build a graph between all corresponding image regions, from which we extract clusters of related images and repeated elements. In the following, we present briefly each of these steps. 
	
	\subsection{Identifying region correspondences}

	Our discovery procedure for a pair of images is visualized in Figure~\ref{fig:discovery}. We
	start by computing dense correspondences between the two images using our learned feature. These will be quite noisy. 
	We first use Hough voting to identify potential groups of consistent correspondences. As a first approximation, each correspondence votes for a translation and change in scale. We then extract the top $10$ translation candidates, and, using a permissive threshold, focus on the correspondences in each group independently.  Within each group, we use RANSAC
	to recover an affine transformation and the associated inliers. This allows to account for some deformation in the copy process, but also variations in the camera viewpoint with respect to the artwork. 

	\subsection{Scoring correspondences}
	\label{sec:score}
	
	After deformations between image regions are identified, we score the correspondence based both on the quality of the match between the features and geometric criteria. We use the following weighted average of the feature similarity:
	
	\vspace{-6mm}
	\begin{eqnarray}
	S(\mathcal{I}) = \dfrac{1}{N} \sum_{i \in \mathcal{I}} e^{(-\dfrac{e_i^2}{2\sigma^2}) } s_i
	\label{eqn:score}
	\end{eqnarray}
	\vspace{-4mm}
	
	where $\mathcal{I}$ is the index of the inlier correspondences, $e_i$ is the error between correspondence $i$ and the geometric model, $s_i$ the similarity of the associated descriptors and $\dfrac{1}{N}$ is  normalization by the number of features in the source image.
	
	\subsection{Correspondence graph}
	
	Using the score $S$, we can separate our dataset into clusters of connected images. These clusters are already interesting and visually appealing, especially for dataset with few repeated details. However, to avoid obtaining very large clusters when many details are repeated in overlapping sets of images, one needs to individually identify each detail region. To do that we built a graph from all the connected image pairs. The nodes are the regions that are mapped between pairs of images. Each image can contain several overlapping regions. We connect regions that are matched to each other as well as regions in the same image that overlap with an Intersection over Union (IoU) score greater than a given threshold (0.5 in our experiments). Finally, we extract the connected components in this graph. Each of them corresponds to a different detail that is repeated in all images of the group.
	
	


	
	
	\section{Experiments}
	
	In this section, we analyse and evaluate our approach. We first present the main datasets we used, including our new annotations of the Brueghel dataset~\cite{brueghel} specifically targeted toward the new task we propose. Second, we provide detailed results and analysis for the task of one-shot visual pattern detection. Finally, we present quantitative and qualitative results for our discovery procedure.
	
	\subsection{Datasets}

	\paragraph{Brueghel.} We introduce new annotations for the Brueghel dataset~\cite{brueghel}, which are available on our project webpage. Indeed, to the best of our knowledge, no other annotation for the task of near duplicate detection in artwork is currently available.
	
\begin{figure*}[!t]
		\vspace{-2mm}
		\begin{center}
			\begin{minipage}{0.48\textwidth}
				\centering    
				\includegraphics[height=10cm]{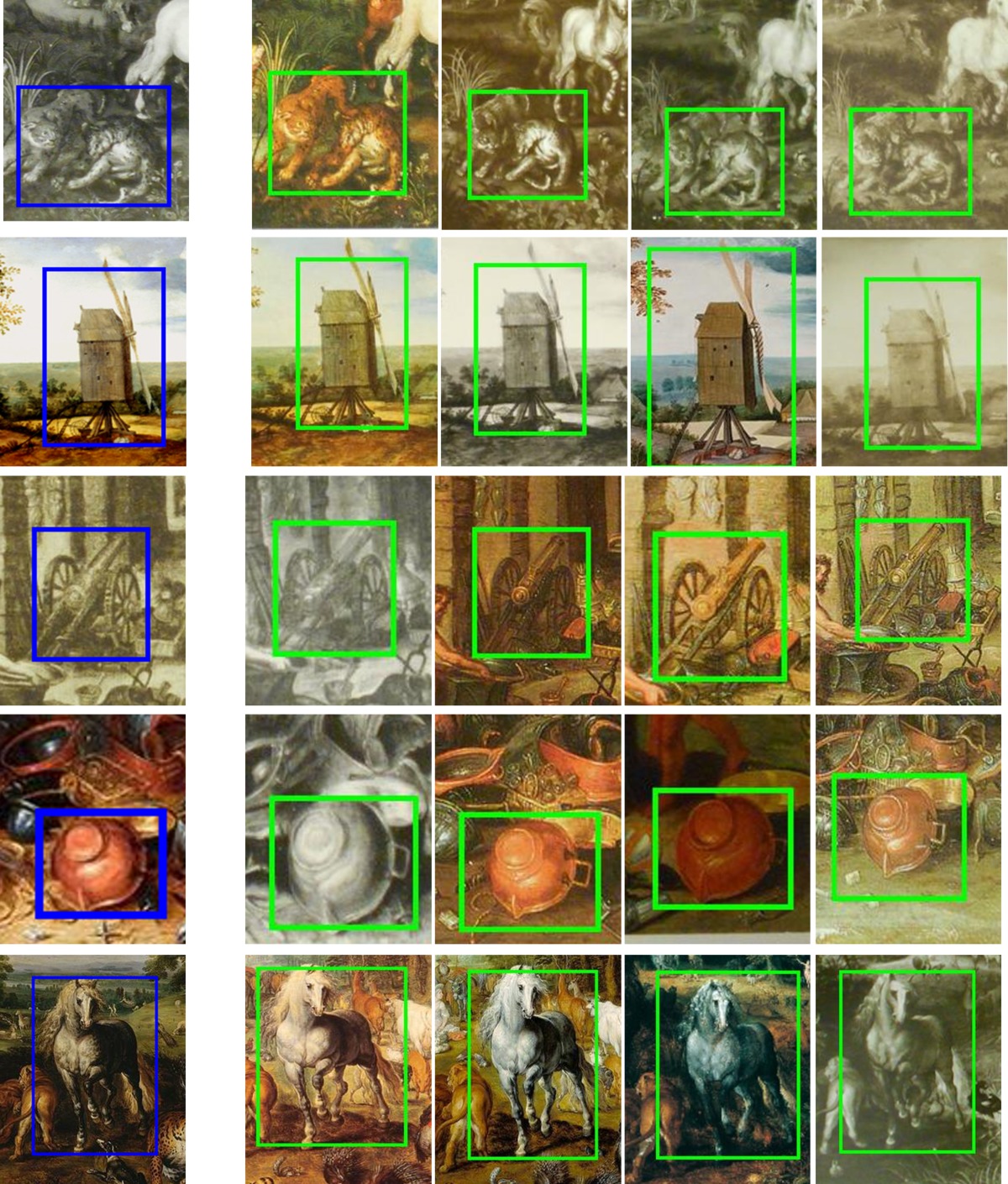}
			\end{minipage}
			\hspace{0.01cm}
			\begin{minipage}{0.48\textwidth}
				\centering
				\includegraphics[height=10cm]{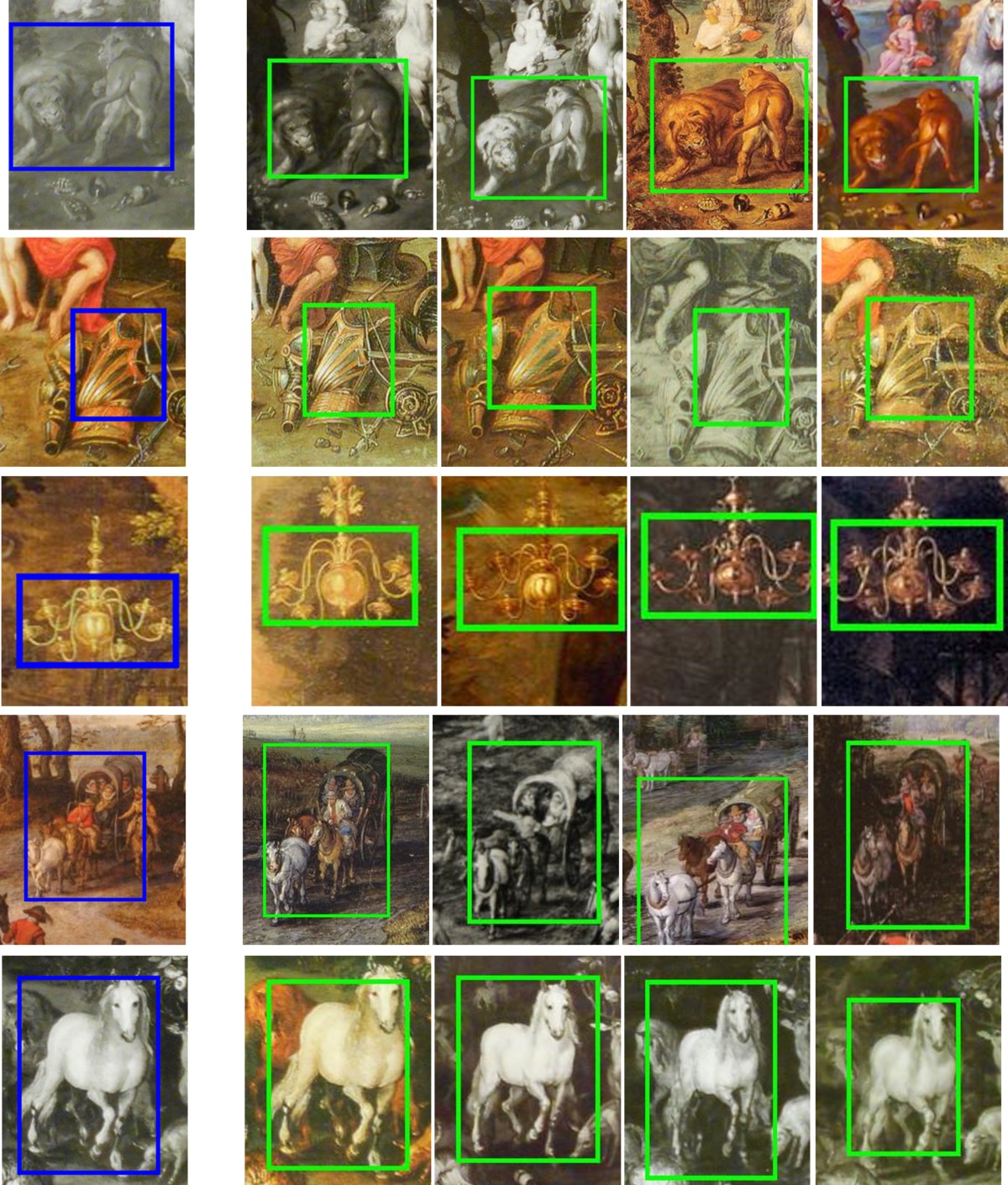}
			\end{minipage}
		\end{center}
		\vspace{-6mm}
		\caption{Detection example with our trained features on the Brueghel dataset. We show the top 4 matches (in green) for one example of query from each of our 10 annotated categories. Notice how the matches style can be different from the one of the query. }
		\label{fig:detection_res}
		\vspace{-5mm}
	\end{figure*}

	The Brueghel dataset contains 1,587 artworks done in different media (e.g. oil, ink, chalk, watercolour) and on different materials (e.g. paper, panel, copper), describing a wide variety of scenes (e.g. landscape, religious, still life) This dataset is especially adapted for our task since it assembles paintings from artists related to the same workshop, who thus had many interaction with each other, and includes many copies, preparatory drawings, and borrowed details. 
	With the help of our art history collaborators, we selected 10 of the most commonly repeated details in the dataset and annotated the visual patterns in the full dataset using the VGG Image Annotator tool~\cite{dutta2016via}. 
	Exemples of the 10 annotated patterns can be seen in Figure \ref{fig:detection_res} as queries (blue boxes). We were careful to select diverse patterns, and for each of them to annotate only duplicates, and not full object classes. Note for example that for the horses and lion classes, we annotated separately two variants of the details (front and back facing lion, front and right facing horse).
	This resulted in 273 annotated instances, with a minimum of 11 and a maximum of 57 annotations per pattern.
	
	These annotations allow us to evaluate one-shot duplicate detection results. In our evaluation, we use an IoU threshold of 0.3 for positives, because precise annotations of the bounding boxes in different environment is difficult and approximate detections would be sufficient for most applications. In practice, our detected bounding boxes, visualised in Figure~\ref{fig:detection_res} (green boxes) often appear more consistent than the annotations. We compute the Average Precision for each query, average them per class and report class level mean Average Precision (mAP).

	\begin{figure}[ht!]
		
		\begin{minipage}{0.02\textwidth}
			\caption*{(a)}
		\end{minipage}
		\begin{minipage}{0.46\textwidth}
			\includegraphics[width=0.99\textwidth]{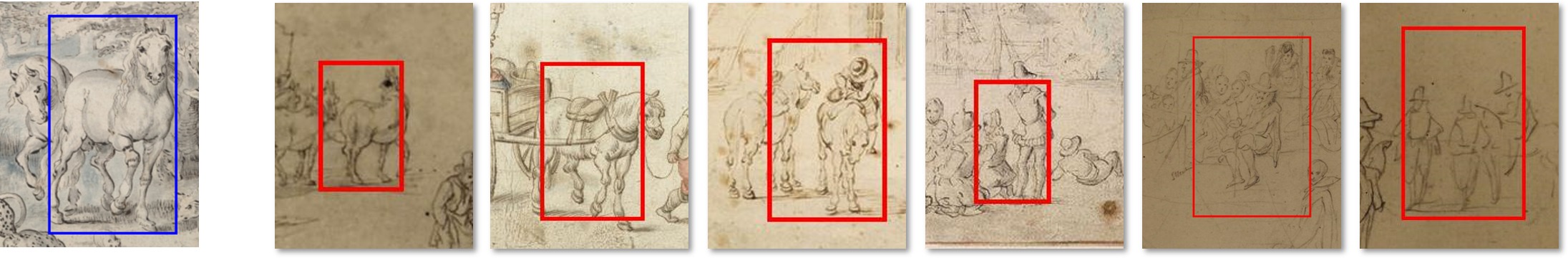}
		\end{minipage}
		
		\begin{minipage}{0.02\textwidth}
			\caption*{(b)}
		\end{minipage}
		\begin{minipage}{0.46\textwidth}
			\includegraphics[width=0.99\textwidth]{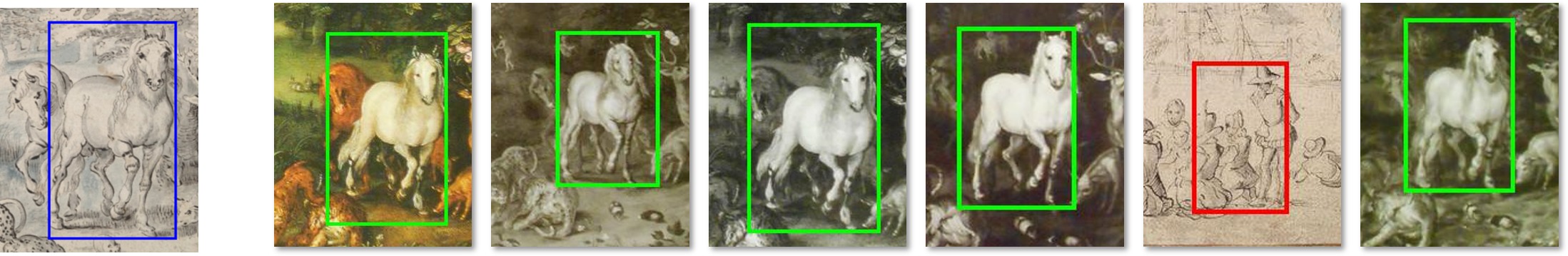}
		\end{minipage}
		
		\begin{minipage}{0.02\textwidth}
			\caption*{(c)}
		\end{minipage}
		\begin{minipage}{0.46\textwidth}
			\includegraphics[width=0.99\textwidth]{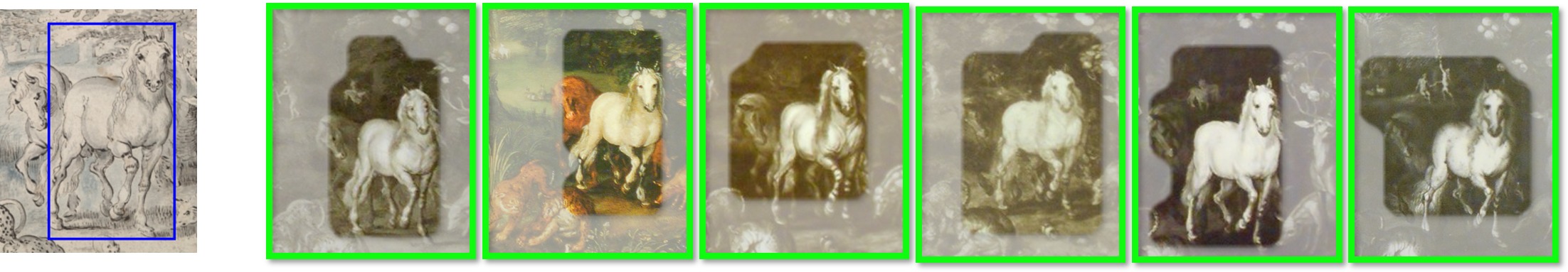}
		\end{minipage}
		
		\vspace{-3mm}
		\caption{From a single query, shown on the left, we show the detection results obtained with cosine similarity with ImageNet feature (a) and our trained features (b) as well as the ones (c) obtained with our features and the discovery score presented in section \ref{sec:score}.}
		\label{fig:comparison}
		\vspace{-3mm}
	\end{figure}
	
	\vspace{-3mm}
	\paragraph{Large Time Lags Locations (LTLL).} While our discovery algorithm targets copy detection in art, it should can detect the same object instances in photographs. We test our algorithm on the LTLL~\cite{fernando2015location} dataset. It contains historic and modern photographs captured from 25 locations over a time interval of more than 150 years. In total the dataset contains 225 historical and 275 modern photographs of the same locations. The task proposed in LTLL is to recognize the location of an old picture using annotated modern photographs. We use our discovery procedure to find the images most similar to the query. As in the original paper~\cite{fernando2015location}, we report classification accuracy. 
	
	\vspace{-3mm}
	\paragraph{Oxford5K.} We also evaluate our approach on Oxford buildings~\cite{philbin2007object} dataset. The dataset contains 5062 images for 11 different landmarks. We follow the standard evaluation protocol and report retrieval mAP for the 55 queries.
	
	\vspace{-3mm}
	\paragraph{DocExplore.} The DocExplore dataset~\cite{en2016new}, which is dedicated to spotting repeated patterns in manuscripts, is the closest existing dataset related to our task and provides extensive comparisons. However the repeated patterns in this dataset are rare and small, all exactly in the same style, with the same colors, and most of the data is text. We thus used it to validate our baseline one-shot detection approach, but could not use it for feature training. DocExplore contains over 1500 images with 1464 instances of 35 different details. For our experiments, we only considered the 18 largest details (the other ones corresponding to small letters).
	
	\vspace{-3mm}
	\paragraph{WikiArt.} To show the generality of our approach, we ran our discovery algorithm on paintings of other artists (Peter Paul Rubens, Dante Gabriel Rossetti and Canaletto) that we collected from WikiArt~\cite{wikiart,wikiart_download} (respectively 387, 195 and 166 artworks).



	\vspace{-1mm}
	\subsection{One-shot detection}
	\vspace{-1mm}
	
	We evaluated our feature learning strategy using one-shot detection. This was performed simply by computing densely features on the dataset and computing their cosine similarity with the features corresponding to the query. The query was resized so its largest dimension in the feature map would be 8. Note that unlike standard deep detection approaches~\cite{girshick2014rich,ren2015faster}, we do not use region proposals because we want to be able to match regions which do not correspond to objects.
	
	Examples results using this approach for each of the 10 details we annotated on the Brueghel dataset are shown in Figure~\ref{fig:detection_res}. It gives a sense of the difficulty of the task we target and the quality of the results we obtain. Note for example how the matches are of different styles, and how the two types of lions (top row) and the two types of horses (bottom row) are differentiated. In the following, we compare these results with baselines and analyse the differences.
	
	\vspace{-3mm}
	\paragraph{Validation on DocExplore.}  To validate that our one-shot detection approach is competitive with classical methods for finding repeated details, 
	we ran it on the DocExplore dataset with ResNet-18 features trained on ImageNet. Our cosine-similarity based dense approach resulted in mAP of 55\% on the 18 categories we considered, a clear improvement compared to the best performance of 40\% obtained in~\cite{en2016new} with classical approaches.


	\begin{table}[h]
		\centering
		\footnotesize
		\vspace{-1mm}
		\begin{tabular}{lcc}
			\hline
			\vspace{-1mm}
			\thead{Feature \textbackslash \ Method} 
			&\thead{ Cosine similarity} & \thead{Discovery score}\\
			\hline
			ImageNet pre-taining & 58.0  & 54.8\\
			Context Prediction~\cite{doersch2015unsupervised}& 58.8  & 64.29\\  
			Ours (trained on Brueghel) & \textbf{75.3} & \textbf{76.4}\\ 
			Ours (trained on LTLL) & 65.2 & 69.95 \\
			\hline
		\end{tabular}
		\vspace{-3mm}
		\caption{Experimental results on Brueghel, IoU $>$ 0.3.}
		\label{tab:detection_brueghel}
	\end{table}

	\begin{figure}[t!]
		\vspace{-2mm}
		\begin{minipage}{0.04\textwidth}
			
		\end{minipage}
		~
		\begin{minipage}{0.12\textwidth}
			\hspace{10mm}\scriptsize{Iter 1800}
		\end{minipage}
		~
		\begin{minipage}{0.12\textwidth}
			\hspace{10mm}\scriptsize{Iter 3600}
		\end{minipage}
		~
		\begin{minipage}{0.12\textwidth}
			\hspace{10mm}\scriptsize{Iter 5400}
		\end{minipage}
	
		\begin{minipage}{0.04\textwidth}
			\centering
			\scriptsize{Source \\ Image}
		\end{minipage}
		~
		\begin{minipage}{0.12\textwidth}
			\centering
			\includegraphics[width=\linewidth, height=2cm, keepaspectratio]{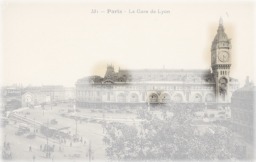}
		\end{minipage}
		~
		\begin{minipage}{0.12\textwidth}
			\centering
			\includegraphics[width=\linewidth, height=2cm, keepaspectratio]{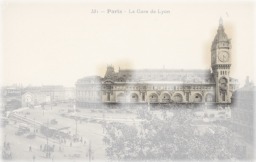}
		\end{minipage}
		~
		\begin{minipage}{0.12\textwidth}
			\centering
			\includegraphics[width=\linewidth, height=2cm, keepaspectratio]{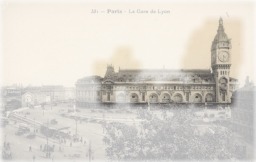}
		\end{minipage}

		\begin{minipage}{0.04\textwidth}
			\centering
			\scriptsize{Target \\ Image}
		\end{minipage}
		~
		\begin{minipage}{0.12\textwidth}
			\centering
			\includegraphics[width=\linewidth, height=2cm, keepaspectratio]{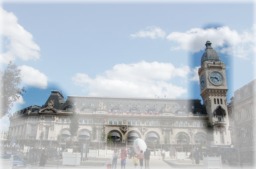}
		\end{minipage}
		~
		\begin{minipage}{0.12\textwidth}
			\centering
			\includegraphics[width=\linewidth, height=2cm, keepaspectratio]{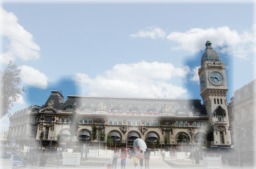}
		\end{minipage}
		~
		\begin{minipage}{0.12\textwidth}
			\centering
			\includegraphics[width=\linewidth, height=2cm, keepaspectratio]{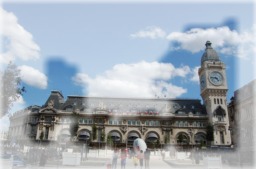}
		\end{minipage}
		\vspace{-3mm}
		\caption{Discovery between images during training.}
		\label{fig:co-dis}
		\vspace{-3mm}
	\end{figure}
	
	\vspace{-3mm}
	\paragraph{Comparison and analysis on Brueghel.} 
	
	Here, we compare the one shot detection performance with different features using cosine similarity and the score described in equation \ref{eqn:score}. In Figure~\ref{fig:comparison}, we present the top 4 matches from the same query using different approaches. On this example, it can be seen that while ImageNet feature only gets the matches in similar styles, our trained feature obtains duplicated horses in different styles, showing that the learned feature is more invariant to style. Moreover, the matching can still be improved with the discovery score. The corresponding quantitative results are presented in Table \ref{tab:detection_brueghel} and confirm these observations. Indeed learning features improves the score by 17\%. The discovery procedure and score provides a small additional boost, which  is a first validation of our discovery procedure. We also report results for two baselines that we re-implemented: the classical Video Google~\cite{sivic2003video} approach and deep feature learnt with Context Prediction~\cite{doersch2015unsupervised}. With the Video Google~\cite{sivic2003video} baseline, we obtained only 21.53\% as retrieval mAP, showing the difficulty to address our task with SIFT~\cite{lowe2004distinctive} features. 
	For Context Prediction, we trained the network using the Brueghel dataset and the same ResNet18 architecture and ImageNet initialization as for our method. We only obtain an improvement of 0.8\% compared to ImageNet feature, much lower than the 17\% provided by our method. Interestingly, training our feature on the LTLL dataset also gave a boost in performance compared to the ImageNet feature, but is clearly worst than training on the Brueghel data, showing the dataset specific nature of our training.
	
	\vspace{-1mm}
	\subsection{Visual Pattern Discovery}
	\vspace{-1mm}
	\begin{figure*}[!t]
		\centering
		\vspace{-6mm}    
		\includegraphics[width=0.97\linewidth]{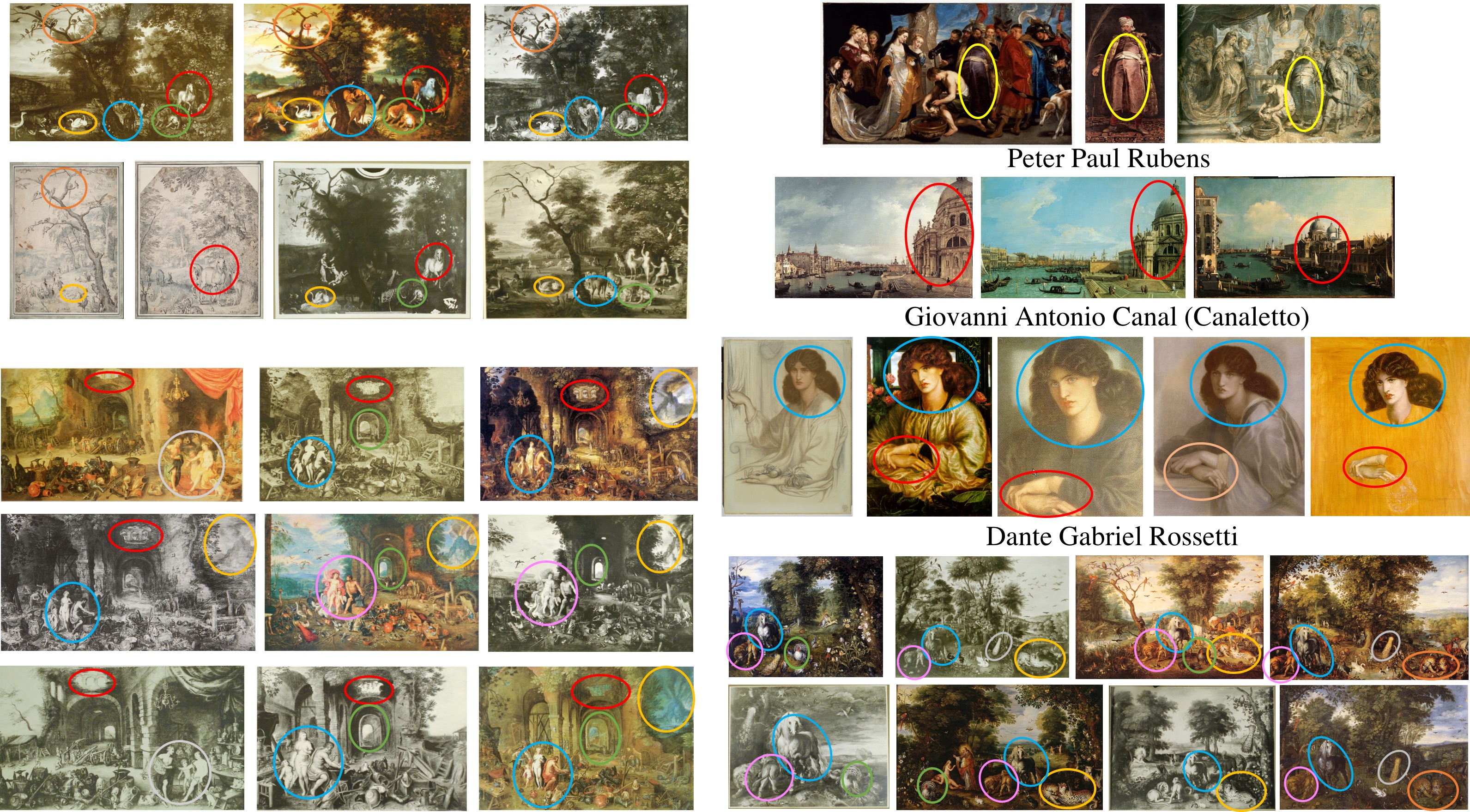}
		\vspace{-3mm}    
		\caption{Example of image clusters discovered by our method. The three cluster without artist names correspond to data from the Brueghel dataset \cite{brueghel}}
		\label{fig:disco}
		\vspace{-5mm}
	\end{figure*}
	
	
	
	In this part, we focus more specifically on our discovery procedure, show qualitative results on various datasets, and evaluate quantitatively for place recognition on the LTLL dataset.
	
	\vspace{-3mm}
	\paragraph{Training visualization.} To visualise the influence of the feature training for our discovery task, we selected a pairs of matching images and ran discovery on them with the features at different steps of training. Figure \ref{fig:co-dis} visualises the results on a pair of images from the LTLL dataset. During training, larger and larger parts of the images can be matched in a consistent way, and be discovered as similar elements by our method. This shows both the efficiency of our feature training and its relevance for our task.

	\begin{table}[t]
		\centering
		\footnotesize
		\vspace{-1mm}
		\begin{tabular}{lcc}
			\hline
			\vspace{-1mm}
			\thead{Method }    & \thead{LTLL (\%)} & \thead{Oxford (\%)}\\ \hline
			B. Fernando et al.\cite{fernando2015location}  & 56.1 & -\\
			F. Radenovi{\'c} et al.\cite{radenovic2017fine}  & -  & 87.8\\
			ResNet18 max-pool, image level & 59.8 & 14.0\\
			ResNet18 + discovery & 80.9  & 85.0 \\  
			Ours (trained LTLL + discovery) & \textbf{88.5} & 83.6\\
			Ours (trained Oxford + discovery) & 85.6 & \textbf{85.7}\\
			\hline
		\end{tabular}
		\vspace{-3mm}
		\caption{Classification accuracy on LTLL and retrieval mAP on Oxford5K}
		\label{tab:disc-LTLL}
		\vspace{-3mm}
	\end{table}
	
	\vspace{-3mm}
	\paragraph{Quantitative analysis on one-shot localization.}
	We evaluate our approach on one-shot localization for both the LTLL and Oxford5K datasets. The results are reported in Table~\ref{tab:disc-LTLL}. We compare our discovery score to cosine similarity with standard max-pooled features as well as the state of the art results of~\cite{fernando2015location} on LTLL and~\cite{radenovic2017fine} on Oxford5K.

	On LTLL, we use the class of the nearest neighbour in modern photographs to localise the historical images. Using the discovery score provides a very important boost compared to the results of~\cite{fernando2015location} and the max-pooled features. Using our fine-tuning procedure on the LTLL dataset improves again the results, demonstrating again the interest of our proposed dataset specific fine-tunning procedure.
	
	
	Similarity, on the Oxford5K dataset, we obtain an important boost using the discovery score compared to cosine similarity with max-pooled features. Fine-tuning the features on Oxford5K improves the mAP by 0.7\%. This improvement is less important than on LTLL, which is expected since there is no specific domain gap between queries and targets in the Oxford5K dataset. Our result on Oxford5K is also comparable to the state-of-the-art result obtained in~\cite{radenovic2017fine} which performs fine-tuning on a large image collection with ResNet101. As expected the retrieval mAP is better when fine-tuning on the Oxford dataset than on LTLL.
	
	\vspace{-3mm}
	\paragraph{Qualitative results.} We show example of our discovery results in Figure \ref{fig:disco}. More results are available in our project webpage.

	\begin{figure}[h!]
		\vspace{-3mm}
		\begin{minipage}{0.5\textwidth}
			\begin{center}
				\includegraphics[width=\linewidth, height=2cm, keepaspectratio]{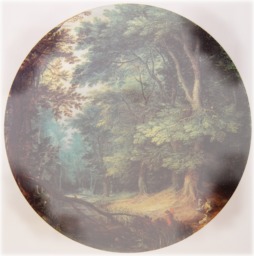}
				\includegraphics[width=\linewidth, height=2cm, keepaspectratio]{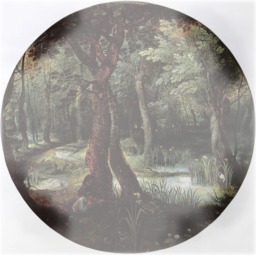}
				\includegraphics[width=\linewidth, height=2cm, keepaspectratio]{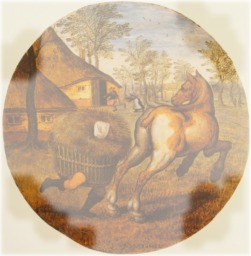}
				\includegraphics[width=\linewidth, height=2cm, keepaspectratio]{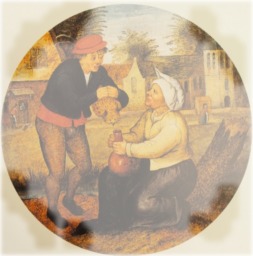}
			\end{center}
		\end{minipage}
		
		\begin{minipage}{0.5\textwidth}
			\begin{center}
				\includegraphics[width=\linewidth, height=2cm, keepaspectratio]{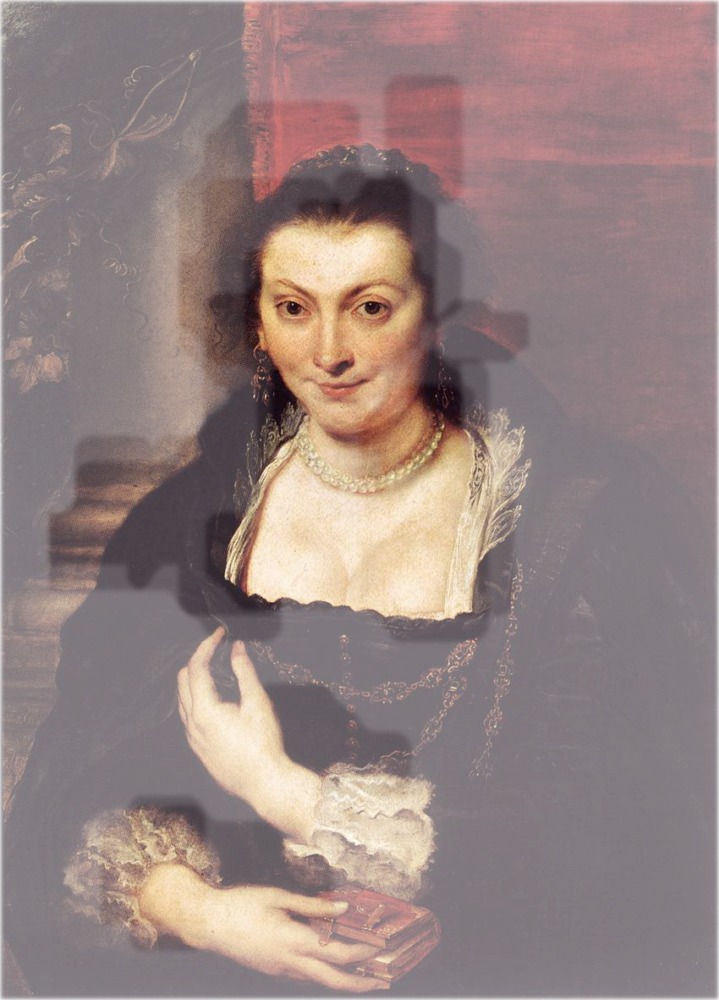}
				\hspace{4mm}
				\includegraphics[width=\linewidth, height=2cm, keepaspectratio]{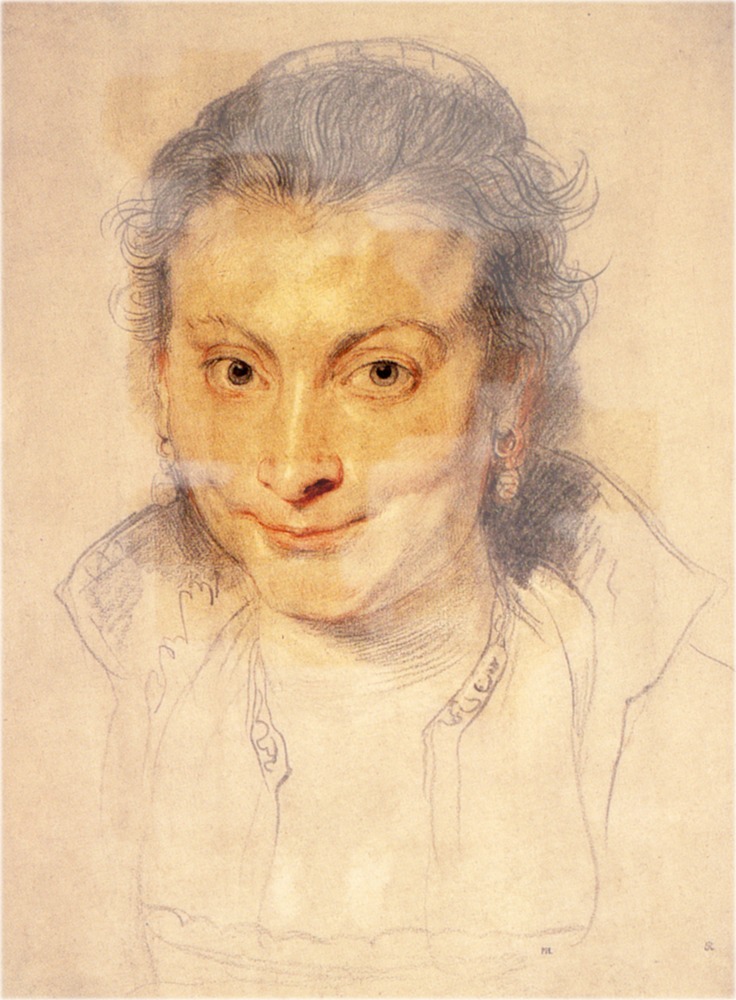}
				\hspace{4mm}
				\includegraphics[width=\linewidth, height=2cm, keepaspectratio]{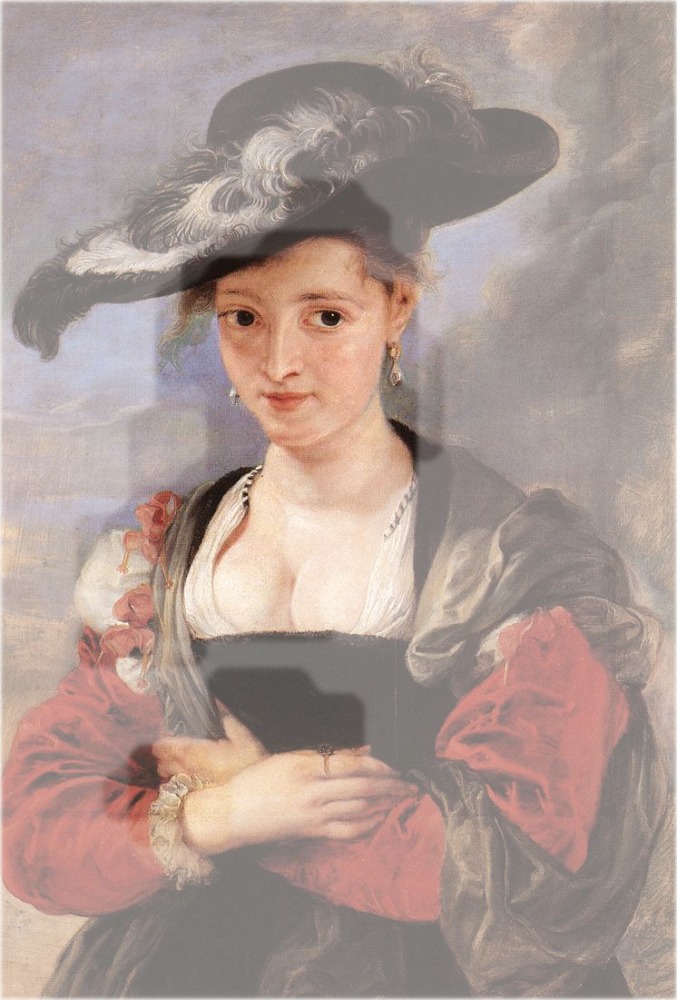}
				\hspace{4mm}
				\includegraphics[width=\linewidth, height=2cm,
				keepaspectratio]{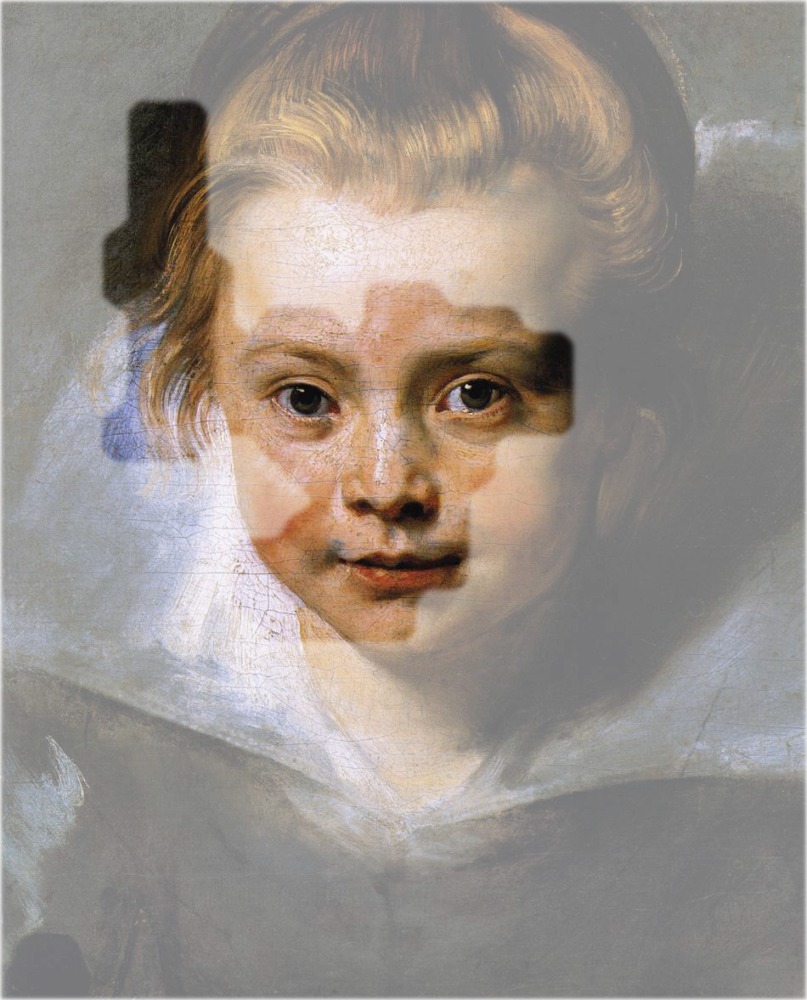}
			\end{center}
		\end{minipage}
		\vspace{-3mm}
		\caption{Failure examples in paintings of the Brughel dataset and from Peter Paul Rubens. }
		\vspace{-3mm}
		\label{fig:failure}
	\end{figure}

	\section{Limitations and Discussion}
	
	When performing discovery on different datasets, we observed some interesting failure modes visualized in Figure~\ref{fig:failure}. In the Brughel dataset, we discovered the identical circular frame of a set of paintings as a repeated pattern, as well as matched the faces in a set of similar but not identical paintings from portrait collections of Peter Paul Rubens.
	
	More generally our method has several limitations. First, as discussed in Appendix~\ref{sec:computational_cost}, the time to perform discovery is important, 
	which limits the size of the datasets we can handle to a few thousands images. Second, our discovery procedure relies on an affine transformation model, which might not always be rich enough. Finally, our feature learning requires having access to a dataset which includes many repeated patterns and a good feature initialization.
	
	\section{Conclusion}

	We have introduced a new approach to adapt features for instance matching on a specific dataset without human supervision. We have demonstrated quantitatively the promise of our method both for one-shot cross-modal detection and for cross-modal instance discovery. Last but not least, we demonstrate diverse near duplicate discovery results in several artwork datasets.
	
	\vspace{-5mm}	
	\paragraph{Acknowledgments} This work was partly supported by ANR project EnHerit ANR-17-CE23-0008, project Rapid Tabasco, NSF IIS-1633310, Berkeley-France funding, and gifts from Adobe to Ecole des Ponts. We thank Shiry Ginosar for her advice and assistance, and Xiaolong Wang, Shell Hu, Minsu Cho, Pascal Monasse and Renaud Marlet for fruitful discussions, and Kenny Oh, Davienne Shields and Elizabeth Alice Honig for thier help on defining the task and building the Brueghel dataset.

	{\small
		\bibliographystyle{ieee}
		\bibliography{egbib}
	}

	
	\appendix
	\section*{Appendix}
	\addcontentsline{toc}{section}{Appendix}
	\renewcommand{\thesubsection}{\Alph{subsection}}
	\subsection{Positive region configuration}
	\label{sec:exp_diff_region}
	
\begin{figure}[h!]
		\centering    
		\includegraphics[width=6cm, height=3.5cm]{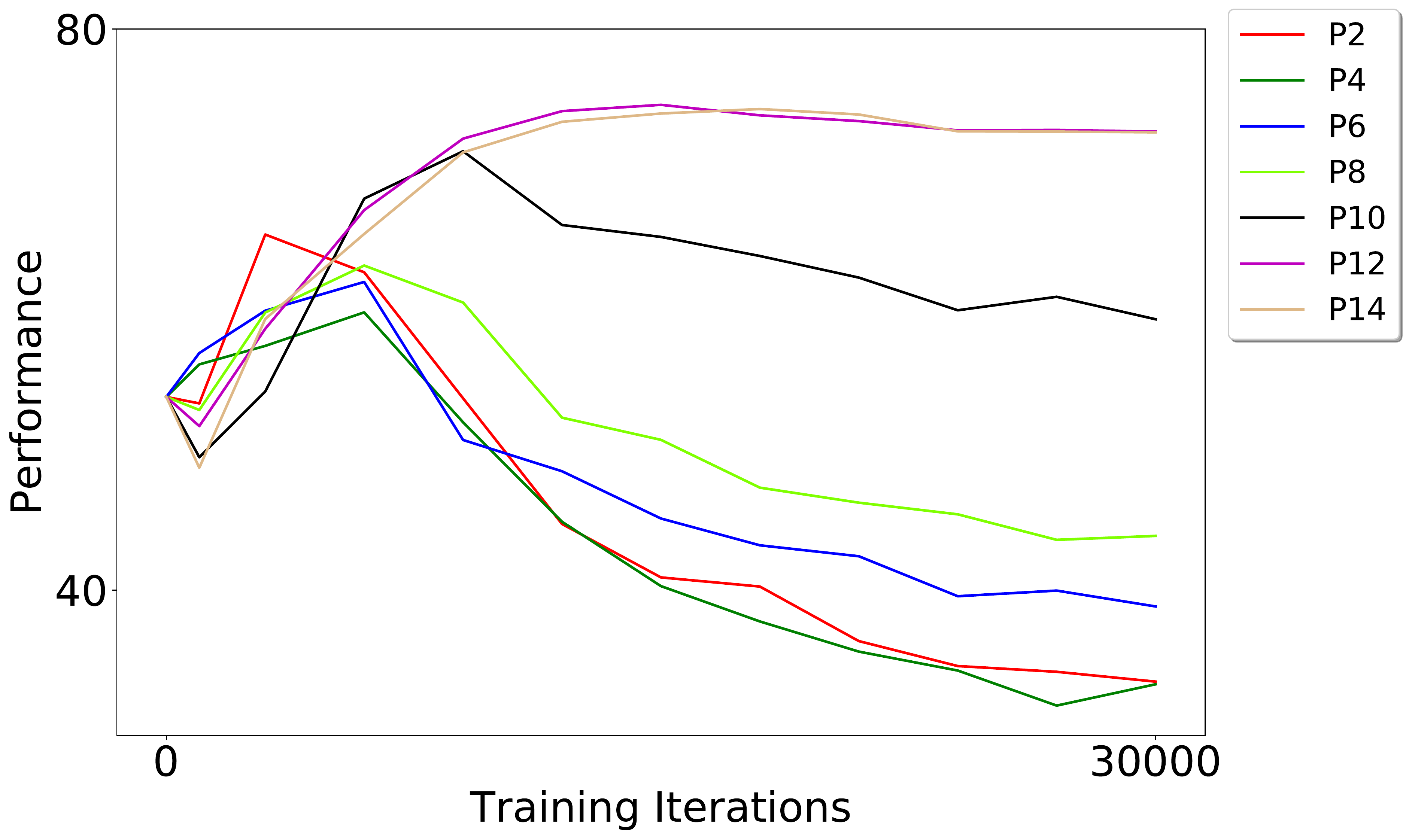}
		\vspace{-4mm}
		\caption{Evolution of the mean Average Precision for one-shot matching on the Brueghel dataset during training. Performance decreases after a few iterations for settings where we extract positive regions correlated with the proposal region.}
		\label{fig:brueghel_training}
	\end{figure}

	We now focus on evaluating the different positive region settings described in Section~\ref{sec:mpp} and Figure~\ref{fig:train_setting}. For each of them, we analyse the performance of the features on one-shot learning on the Brueghel dataset and its evolution during training. The results can be seen in Figure~\ref{fig:brueghel_training}. Interestingly, the performance initially always improves over ImageNet features. However, when the positive region is close to the proposal region, the performance decreases after some iterations of our training procedure, and ends up with worse performance than the initial features. But if the positive region is far enough from the query (P12 and P14), the performance improves much more and does not subsequently deteriorate. We thus use P12 for all our experiments.
	
	\subsection{Computational cost}
	\label{sec:computational_cost}
	We implement an efficient algorithm for the single shot detection by considering query features as convolutional kernels. We can thus match one query to 53 images/sec on a single GPU (Geforce GTX 1080 Ti). The discovery step is slower, since it requires matching all the features of one image to another, and takes approximately 0.2 seconds/pair of images on a single GPU. It takes about 40 minutes to query one image on Oxford5K using discovery procedure on one GPU and the discovery on the whole dataset of Brueghel took approximately 20 hours using 4 GPUs.

\end{document}